\documentclass[nofootinbib,superscriptaddress,a4paper,twocolumn,longbibliography]{revtex4-1}

\usepackage{graphicx}
\usepackage{dcolumn}
\usepackage{bm}
\usepackage{subfigure}
\usepackage{tikz}
\usepackage{booktabs}

\usepackage[export]{adjustbox}
\newsavebox{\imagebox}

\usepackage{hyperref}

\graphicspath{
	{pics/}
}
\begin{document}

\title{Scientific intuition inspired by machine learning generated hypotheses}
\author{Pascal Friederich}
\email[]{pascal.friederich@kit.edu}
\affiliation{Chemical Physics Theory Group, Department of Chemistry, University of Toronto, Canada.}
\affiliation{Department of Computer Science, University of Toronto, Canada.}
\affiliation{Institute of Theoretical Informatics, Karlsruhe Institute of Technology, Am Fasanengarten 5, 76131 Karlsruhe, Germany.}
\affiliation{Institute of Nanotechnology, Karlsruhe Institute of Technology, Hermann-von-Helmholtz-Platz 1, 76344 Eggenstein-Leopoldshafen, Germany.}

\author{Mario Krenn}
\affiliation{Chemical Physics Theory Group, Department of Chemistry, University of Toronto, Canada.}
\affiliation{Department of Computer Science, University of Toronto, Canada.}
\affiliation{Vector Institute for Artificial Intelligence, Toronto, Canada.}

\author{Isaac Tamblyn}
\affiliation{National Research Council of Canada, Ottawa, Canada.}
\affiliation{Vector Institute for Artificial Intelligence, Toronto, Canada.}

\author{Alán Aspuru-Guzik}
\email[]{alan@aspuru.com}
\affiliation{Chemical Physics Theory Group, Department of Chemistry, University of Toronto, Canada.}
\affiliation{Department of Computer Science, University of Toronto, Canada.}
\affiliation{Vector Institute for Artificial Intelligence, Toronto, Canada.}
\affiliation{Canadian  Institute  for  Advanced  Research  (CIFAR)  Lebovic  Fellow,  Toronto,  Canada}

\date{\today}

\begin{abstract}
\textbf{Abstract}
Machine learning with application to questions in the physical sciences has become a widely used tool, successfully applied to classification, regression and optimization tasks in many areas. Research focus mostly lies in improving the accuracy of the machine learning models in numerical predictions, while scientific understanding is still almost exclusively generated by human researchers analysing numerical results and drawing conclusions. In this work, we shift the focus on the insights and the knowledge obtained by the machine learning models themselves. In particular, we study how it can be extracted and used to inspire human scientists to increase their intuitions and understanding of natural systems. We apply gradient boosting in decision trees to extract human interpretable insights from big data sets from chemistry and physics. In chemistry, we not only rediscover widely know rules of thumb but also find new interesting motifs that tell us how to control solubility and energy levels of organic molecules. At the same time, in quantum physics, we gain new understanding on experiments for quantum entanglement. The ability to go beyond numerics and to enter the realm of scientific insight and hypothesis generation opens the door to use machine learning to accelerate the discovery of conceptual understanding in some of the most challenging domains of science.
\end{abstract}

\maketitle 
\section{Introduction}\label{chp:intro}
Machine learning (ML) recently became a widely used tool with many applications in the physical sciences \cite{carleo2019machine}, ranging from chemistry (for example, prediction of quantum chemistry properties \cite{ramakrishnan2015big}, solving Schr{\"o}dinger's equation \cite{hermann2020deep}, predicting reactions \cite{schwaller2019molecular}, materials discovery \cite{li2020robot} or inverse materials design \cite{gromski2019explore, sanchez2018inverse}) to physics (for example, identification of phases of matter \cite{carrasquilla2017machine}, astronomical object recognition \cite{hezaveh2017fast}, or validation of quantum experiments \cite{agresti2019pattern}) and biology (for example, prediction of protein structures \cite{senior2020improved} or drug design \cite{zhavoronkov2019deep,stokes2020deep}).
Some open challenges regarding the application of machine learning models in natural sciences include the accessibility, homogeneity, amount and quality of available data, as well as a lack of machine learning models which inherently include physical laws, limiting the interpretability of the models’ predictions.
While ML models are successfully used and optimized to accelerate numerical predictions or to recognize or generate patterns in existing data, it is rarely inquired how the machine finds solutions, {\it i.e.} which patterns and correlations it detected and exploited. Thus, the scientific insight obtained by the model is not directly transferred to human scientists.
First attempts to use artificial intelligence in physical sciences aimed to directly answer scientific questions, {\it e.g.} determine the location of protein encodings in the genome \cite{king2009automation}.
Further attempts to employ machine learning models to obtain insight and help scientists to develop theories were focused on rediscovering solutions to already solved problems, {\it e.g.} to rediscover the coordinate transformation in astrophysical \cite{iten2020discovering} and nonlinear dynamical systems \cite{lusch2018deep}, or to detect symmetries and conservation laws \cite{Wetzel2020discovering}.
The methods used in these cases enforce information bottlenecks or interpretable transformations in the ML model that then can inspire scientific understanding \cite{roscher2020explainable}. However, to our knowledge such methods were mostly applied to solved problems and have not been used yet to obtain novel insight and answers to questions that are not well understood yet.\\

In this work, we propose to use machine learning and systematic data analysis to automate further the process of generation of {\it interpretable} scientific hypotheses.
We demonstrate the applicability of the approach using two questions in the natural sciences - a rediscovery task of chemistry knowledge (hydrophobicity and molecular energy levels in simple as well as application relevant molecules) and the discovery of new intuitions in physics (quantum optics).
We show that our approach “rediscovers” but also extends known chemical rules of thumb for solubility and energy levels of organic molecules with application in organic photovoltaics and organic light-emitting diodes and helps us to better understand the entanglement created in quantum optical experiments.

Our model represents its findings in a graph representation which is directly related to chemical or physical instances in the specific scientific domain. The results are statements regarding distinct subgraphs that can easily be comprehended and therefore, scientifically interpreted and understood by experts. This is in stark contrast to conventional machine learning models where the internal representations are only indirectly connected with the real physical entities and thus hard to impossible to interpret.

\begin{figure*}[hbt!]
    \centering
    \includegraphics[width=0.9\textwidth]{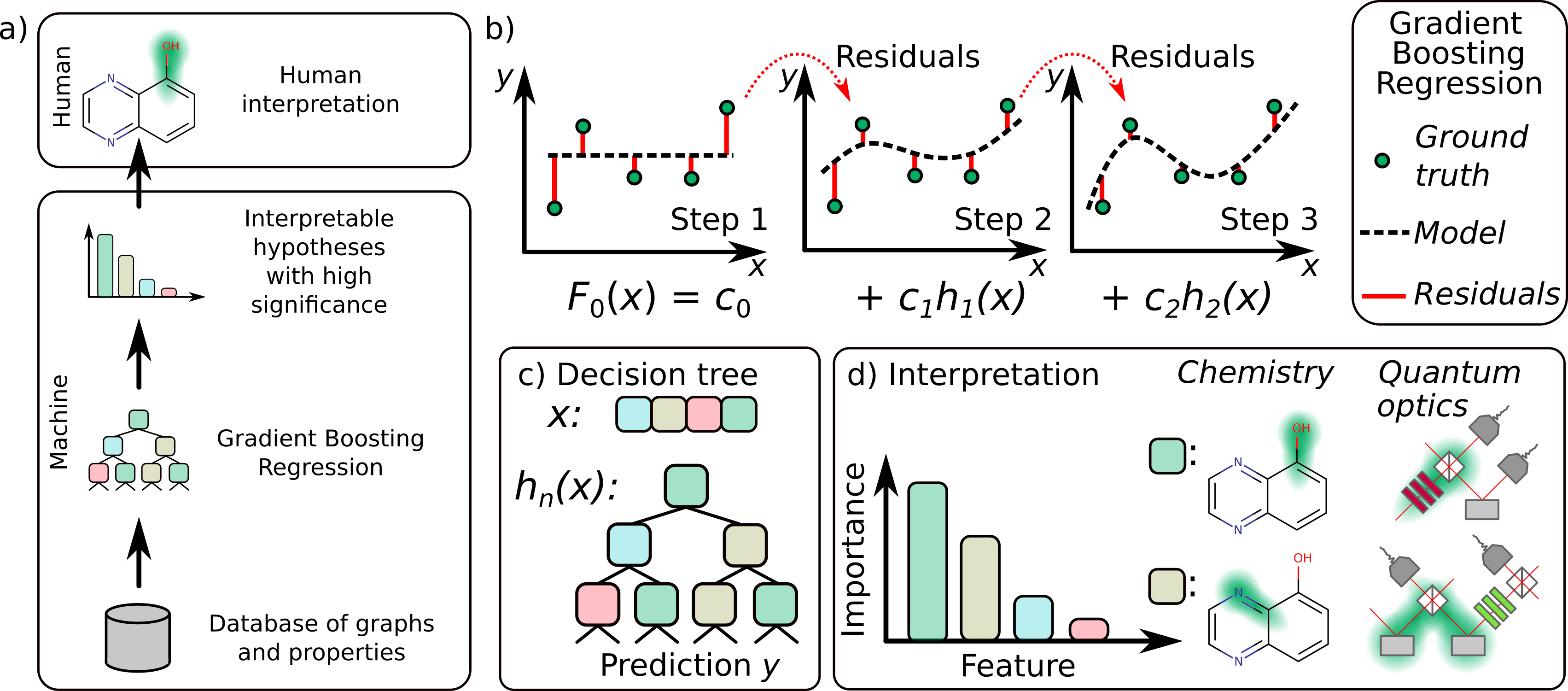}
    \caption{\textbf{Workflow for automated hypothesis generation.} a) General workflow, starting with a database of graphs and respective properties, followed by training of a machine learning model that allows for the extraction of feature importances, {\it e.g.} Gradient Boosting Regression. Features with high importance are combined and analysed in a way that facilitates interpretation by researchers in order to stimulate scientific insight. b) Schematic illustration of the Gradient Boosting Regression method \cite{friedman2001greedy}, where multiple simple decision tree models are trained sequentially. Each new decision tree is trained to correct the residual errors (red lines) of the previous models, so the final prediction $F_0(x)$ can be written as a sum of the mean label $c_0$ and a weighted series of models $h_i(x)$, where each $h_i$ predicts the deviation of the previous $i-1$ models from the ground truth. c) Each decision tree is trained on samples that are represented using predefined input features (coloured squares) and uses their values to split the data set sequentially into smaller subsets which are used for the predictions. The subgraph based input representation used in this work allows a direct interpretation of the feature importances (d) that are computed based on a quantification of how meaningful features are for the accuracy of the machine learning model.}
    \label{fig:overview}
\end{figure*}

\section{Method}\label{chp:method}
\textbf{Computer generated hypotheses.}
We suggest an automated workflow for ML-based generation of human interpretable scientific hypotheses as illustrated in Figure \ref{fig:overview}a. The workflow is based on a reference database of calculated (potentially also measured) data points with graph-based structure and with corresponding target properties. A binary feature vector describing presence/absence of automatically generated subgraphs \cite{rogers2010extended} is used to train a tree ensemble method, {\it e.g.} Gradient Boosting \cite{friedman2001greedy} or Random Forrest Regression/Classification \cite{ho1995random,breiman2001random}, that allows for the quantification of feature importances. Based on the features with the highest importance, a list of hypotheses is generated. Each hypothesis has the human understandable form\\

\textit{“Feature i leads to an increase/decrease of target property of strength s”}\\

where $i$ is the index of the corresponding feature (subgraph) in the input and strength $s$ quantifies the degree of correlation between feature $i$ and the target property. 
High feature importance does not necessarily correspond to a high direct correlation with the target feature. In many cases, multiple features have to be combined in order to become predictive, even if the single features individually do not help in the predicting the target property. Therefore, important features are combined using logical operations (\textit{and}, \textit{xor}, ...) to automatically generate combined features which, especially in presence of higher-order correlations, can be directly interpreted by researchers.\\

\begin{figure}[hbt!]
    \centering
    \includegraphics[width=0.45\textwidth]{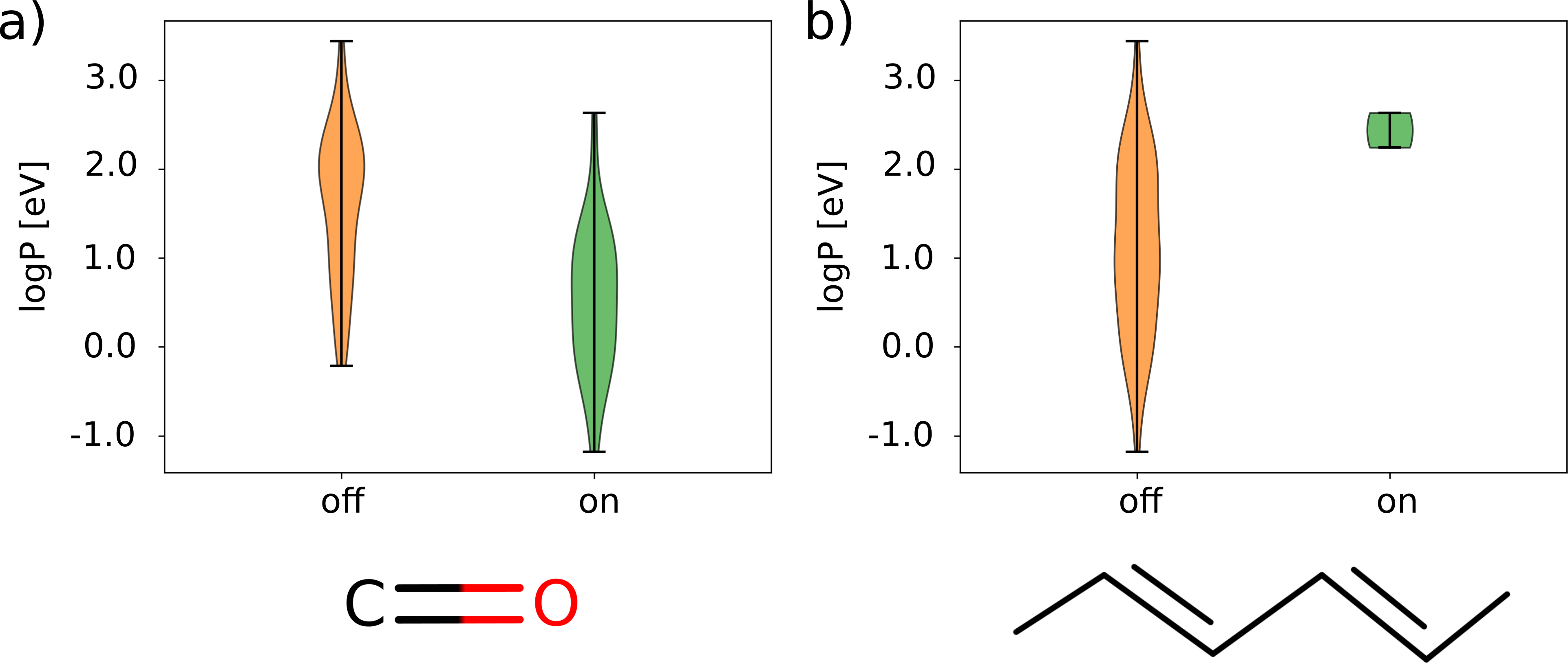}
    \caption{\textbf{Hypotheses about molecular solubility.} a) Lower logP values (better solubility in water compared to octanol) can be achieved using the carbonyl groups, while b) conjugated carbon chains lead to higher logP values.}
    \label{fig:solubility}
\end{figure}

\textbf{Input representation and experiments.} In this work, we test this workflow on two experiments in chemistry and physics. The first experiment targets the automated generation of intuitive rules that determine molecular properties, whereas the second aims at hypothesis generation for entanglement properties of quantum optical experiments. In both cases, we can describe the data points as graphs (molecules and quantum optical experiments), where nodes are chemical elements or optical instruments while edges are chemical bonds or photon paths travelling through the setup. This allows us to use fingerprinting techniques to generate input representations (bit-vectors),  {\it e.g.} using the algorithm for circular extended-connectivity fingerprints \cite{rogers2010extended}. This iterative algorithm generates a unique representation of each node, including its local environment. In each iteration, hashing functions are used to aggregate the information (predefined node and edge features) of the next nearest neighbors of each node, thus implicitly integrating information of one additional neighbor shell in each iteration. In the end, a hashing function is used to map all subgraphs found in the graphs to bit-vectors. Each entry in these bit-vectors encodes the presence or absence of a certain subgraph. A similar approach has been used in Lopez {\it el al.}\cite{lopez2017design} to determine molecular substructures in molecules for organic solar cells that lead to high power conversion efficiencies. Other models that link the presence of subgraphs (or more generally features) in the input data to properties can potentially be employed in our workflow (see e.g. Duvenaud {\it et al.} \cite{duvenaud2015convolutional} where molecular fragments are identified that correlate with toxicity, the Grad-CAM method by Selvaraju {\it et al.} \cite{selvaraju2017grad} for convolutional neural networks or the GNNExplainer by Ying {\it et al.} \cite{ying2019gnnexplainer}). In contrast to this work, some of these approaches depend on the analysis of single samples and thus only indirectly allow to conclude about an entire data set. Furthermore, these approaches assign importance indicators to single nodes or edges of a graph, which are not necessarily binary numbers, which complicates the direct interpretation. Due to their general applicability to all graphs where node and edges can be represented by one or multiple categorical features, we focused on automatically generated circular fingerprints in this work.\\

\section{Results}\label{chp:results}
To test the automated hypothesis generation workflow, we performed experiments in two scientific domains, molecular chemistry (Section \ref{chp:subsecChem}) and quantum optical experiments (Section \ref{chp:subsecPhys}). We computed physical properties of these graphs and used the generated data sets and the workflow described in Figure \ref{fig:overview} to automatically generate hypotheses that can be either compared to a collection of widely known chemical rules of thumb or that can help to better understand entanglement in quantum optical experiments for designing future experiments.\\

\subsection{Chemical intuition for solubility, energy levels}\label{chp:subsecChem}
In case of the chemistry experiment, we used two prototypical target properties - the water-octanol partition coefficient which describes the solubility of molecules in water (polar) vs. octanol (non-polar) as well as the energy of the highest occupied molecular orbital. Both properties are of high relevance for the application of molecules as pharmaceuticals or in electronic devices, {\it e.g.} for organic solar cells, organic light-emitting diode (OLED) displays or organic flow batteries. We furthermore analysed existing application-specific data sets, namely a data set of thermally activated delayed fluorescent (TADF) molecules as emitter molecules for OLEDs \cite{gomez2016design}, the Harvard Clean Energy project data set \cite{hachmann2011harvard,lopez2016harvard} and a data set of non-fullerene acceptor molecules for organic solar cells \cite{lopez2017design}. Solubility and energy levels are relatively well understood and for both properties there exist several widely known rules of thumb, often described as chemical intuition, which describes how certain functional groups influence them. Our experiment aims to test whether the automated hypothesis generation method can “rediscover” those rules and potentially add new or refined rules. For frontier orbital gaps reported in the Harvard Clean Energy data set and the non-fullerene acceptor data set as well as for singlet-triplet energy splittings reported in the TADF data set, there exists less chemical intuition on how to influence and tune them.\\

\begin{figure}[hbt!]
    \centering
    \includegraphics[width=0.37\textwidth]{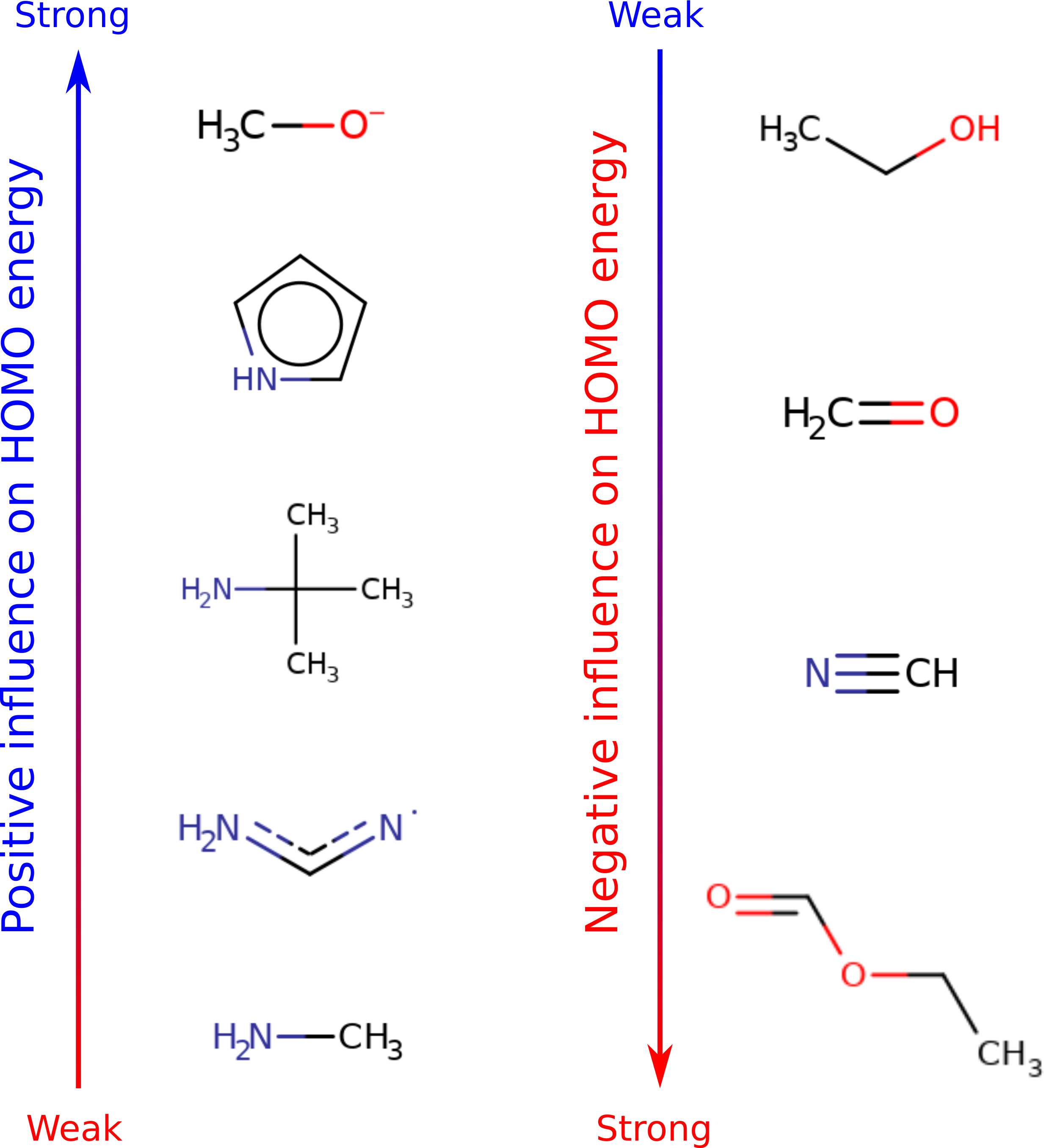}
    \caption{\textbf{Hypotheses about molecular energy levels.} Molecular subgraphs with a positive (left) and negative (right) influence on the HOMO energy. The groups “discovered” by our automated workflow are widely known activating (resonance donating or electron donating) and deactivating groups, such as oxido/amino groups and nitrile groups.}
    \label{fig:homolumo}
\end{figure}

Figure \ref{fig:solubility} shows two solubility related hypotheses that were generated using our workflow. Without prior knowledge, the algorithm predicts two widely known chemical groups/motifs for increasing solubility in polar solvents (carbonyl group in Figure \ref{fig:solubility}a) and to increase solubility in non-polar solvents (conjugated carbon chain in Figure \ref{fig:solubility}b).
Figure \ref{fig:homolumo} shows an overview of molecular subgraphs that positively and negatively influence the HOMO energy of a molecule. To our surprise, five of the nine groups shown in the figure can directly be found in chemistry textbooks or Wikipedia when searching for electrophilic aromatic directing groups which can change the energy levels of molecules through the inductive effect and the mesomeric effect.
Specifically, the oxido (O$^{\--}$) group that shows the strongest positive influence on the HOMO is well known for a strong resonance donating and a strong inductive effect which both leads to an increase in HOMO energy. Furthermore, heterocycles that contain nitrogen, as well as amine (NH2) groups are also known for lifting the HOMO level to higher energies. On the other hand, the nitrile group (C$\equiv$N) is one of the most widely known electron-withdrawing groups that lowers the HOMO energy of molecules due to its resonance withdrawing and inductively withdrawing nature.\\

\begin{figure}[hbt!]
    \centering
    \includegraphics[width=0.35\textwidth]{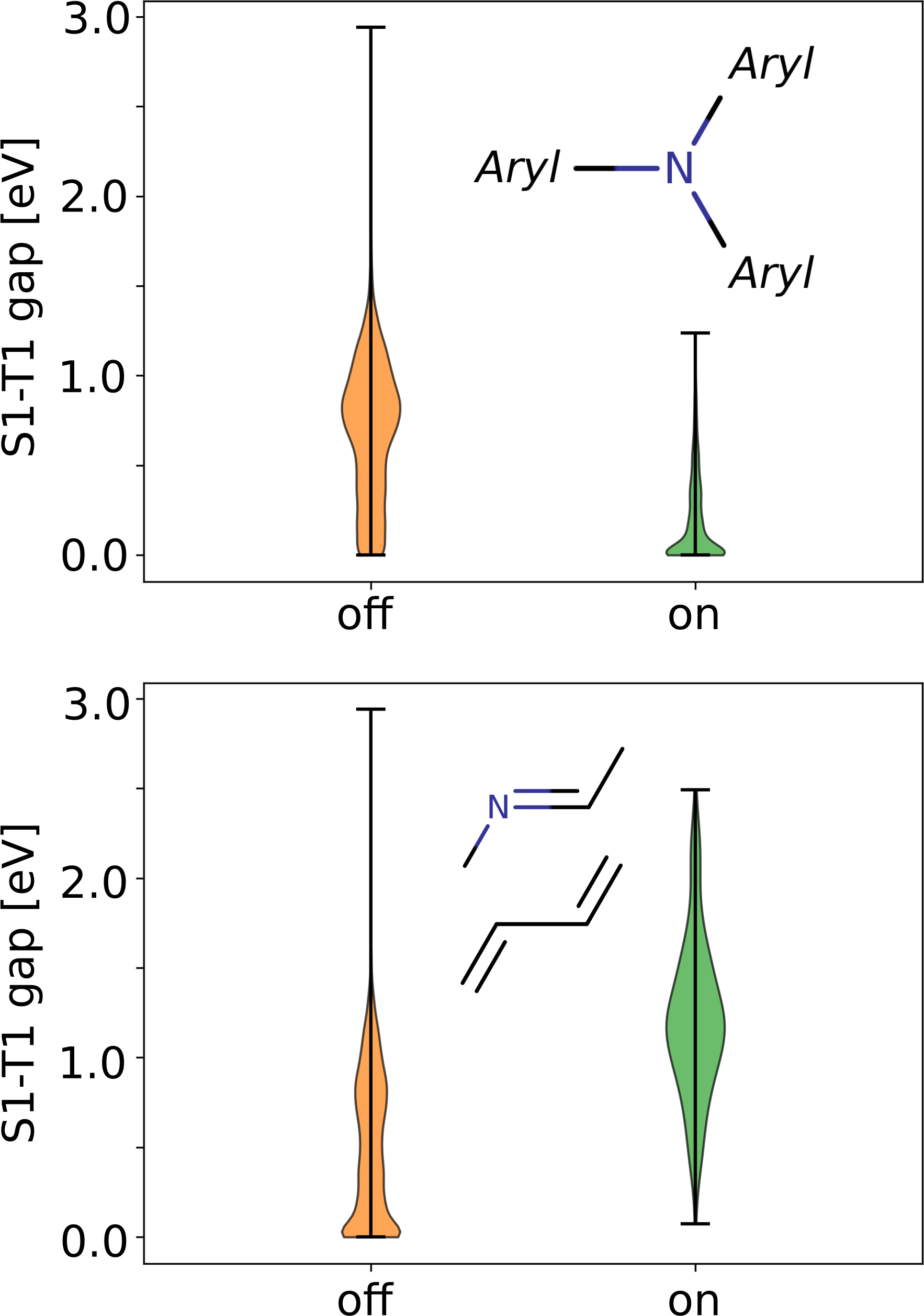}
    \caption{\textbf{Hypotheses singlet-triplet splittings in the TADF data set \cite{gomez2016design}.} The data-driven algorithm finds the well known and widely exploited structure-property relation of triarylamines and small single triplet gaps ($<$0.5 eV, upper panel). However, it finds an additional, less known motif of alternating single-double-bond bridges that are related to increased singlet triplet gaps ($>$0.5 eV, lower panel).}
    \label{fig:TADF}
\end{figure}

\begin{figure*}[hbt!]
    \centering
    \includegraphics[width=0.9\textwidth]{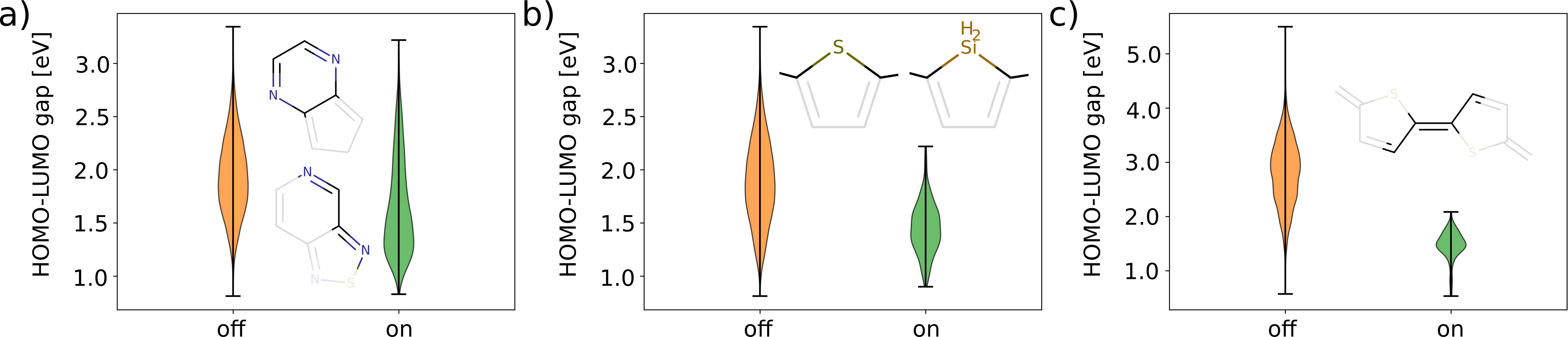}
    \caption{\textbf{Hypotheses about HOMO-LUMO gaps in the Harvard Clean Energy data set \cite{hachmann2011harvard,lopez2016harvard} and a non-fullerene acceptor data set \cite{lopez2017design}.} (a) The automated hypotheses generation protocol rediscovers the widely known relation between extended aromatic systems (containing {\it e.g.} nitrogen heteroatoms) and reduced HOMO-LUMO gaps. (b) Thiophene but also more uncommon silole rings are found to correlate with small HOMO-LUMO gaps. c) Thiophene rings bridged with double bonds (quinoid structures) are found to decrease the HOMO-LUMO gap in the non-fullerene acceptor data set. (Note the different scale in panel (c) compared to (a) and (b), due to differences in the data sets.)}
    \label{fig:homolumogap}
\end{figure*}

The patterns found to be relevant for small HOMO-LUMO gaps in the Harvard Clean Energy data set as well as in the non-fullerene acceptor data set are mostly related to extended aromatic systems and fused aromatic rings (see Figure \ref{fig:homolumogap}a and Figure S1a). This finding is well-understood by chemists due to the widely know relation between the size of an aromatic system ({\it i.e.} the degree of delocalization of $\pi$-electrons) and the frontier orbital gap \cite{gershoni2018predictive}. In the limit of infinite delocalization ({\it e.g.} in graphene), the HOMO-LUMO gap closes completely. This relation was also exploited in the development of conductive polymers, which was awarded with the Nobel Price in Chemistry in 2000 and which created the field of organic electronics \cite{rasmussen20182000}.

However, we additionally found several interesting and surprising patterns both in the photovoltaic data sets (Figure \ref{fig:homolumogap}b/c) and in the TADF dataset (Figure \ref{fig:TADF}). In case of the Harvard Clean Energy data set, we find that aromatic heterocycles with sulfur ({\it e.g.} thiophene rings) as well as silicon heteroatoms ({\it e.g.} silole rings) significantly reduce the HOMO-LUMO gap. While the former are widely used in organic electronics to control energy levels and reduce HOMO-LUMO gaps, silole rings are more unusual.

In the non-fullerene acceptor data set (see Figure \ref{fig:homolumogap}c) we found that thiophene rings connected by double bonds ({\it i.e.} forming a quinoid structure instead of aromatic systems) also significantly reduce the HOMO-LUMO gap, which is a know relation first described by Brédas \cite{bredas1985relationship}. However, such systems require a specific functionalization in the periphery of the molecule to enforce the quinoid structure of the two thiophene rings, which intrinsically is less stable and thus higher in energy than the aromatic structure.

In case of the TADF data set (see Figure \ref{fig:TADF}), we found expected patterns such as triarylamines that correlate with decreased singlet triplet gaps (S1-T1 gaps) as well as rather unexpected patterns ({\it e.g.} conjugated bridges) that are identified by our workflow as chemical groups that highly correlate with large singlet triplet gaps. Low singlet-triplet splittings in TADF molecules are typically achieved by decoupling electron donating and electron accepting parts of a molecule to reduce the exchange interaction between the frontier orbitals which would otherwise lower the triplet state compared to the singlet state and open an undesired singlet-triplet splitting. The decoupling of the fragments can be achieved by introducing twist angles close to 90$^{\circ}$ between the fragments. One way to accomplish this are triarylamines bridges between the fragments. We expect that the conjugated bridges between fragments have precisely the opposite effect: They lead to a planar alignment of the adjacent fragments and thus an enhanced exchange interaction, reduced triplet energies and finally increased singlet-triplet splittings.\\

\subsection{Physical intuitions for quantum experiments}\label{chp:subsecPhys}
As a second example, we use quantum optical experiments for producing high-dimensional, multipartite quantum entanglement \cite{friis2019entanglement,erhard2019advances}. These experiments grow in interest as they allow the investigation of fundamental physical properties - such as local realism \cite{lawrence2017mermin} - in laboratories. Furthermore, such quantum states are the key resources for large and complex quantum communication networks \cite{pivoluska2018layered,hu2020experimental}, which are on the edge of commercial availability. The experimental setups that we consider consist of standard optical components that are used in labs, such as nonlinear crystals for the creation of photon pairs, single-photon detectors, beam splitters, holograms or Dove prisms. Under approximations that are closely resembled in experiments, the final emergent quantum state can be reliably calculated \cite{pan2012multiphoton}.\\

\begin{figure}[hbt!]
    \centering
    \includegraphics[width=0.49\textwidth]{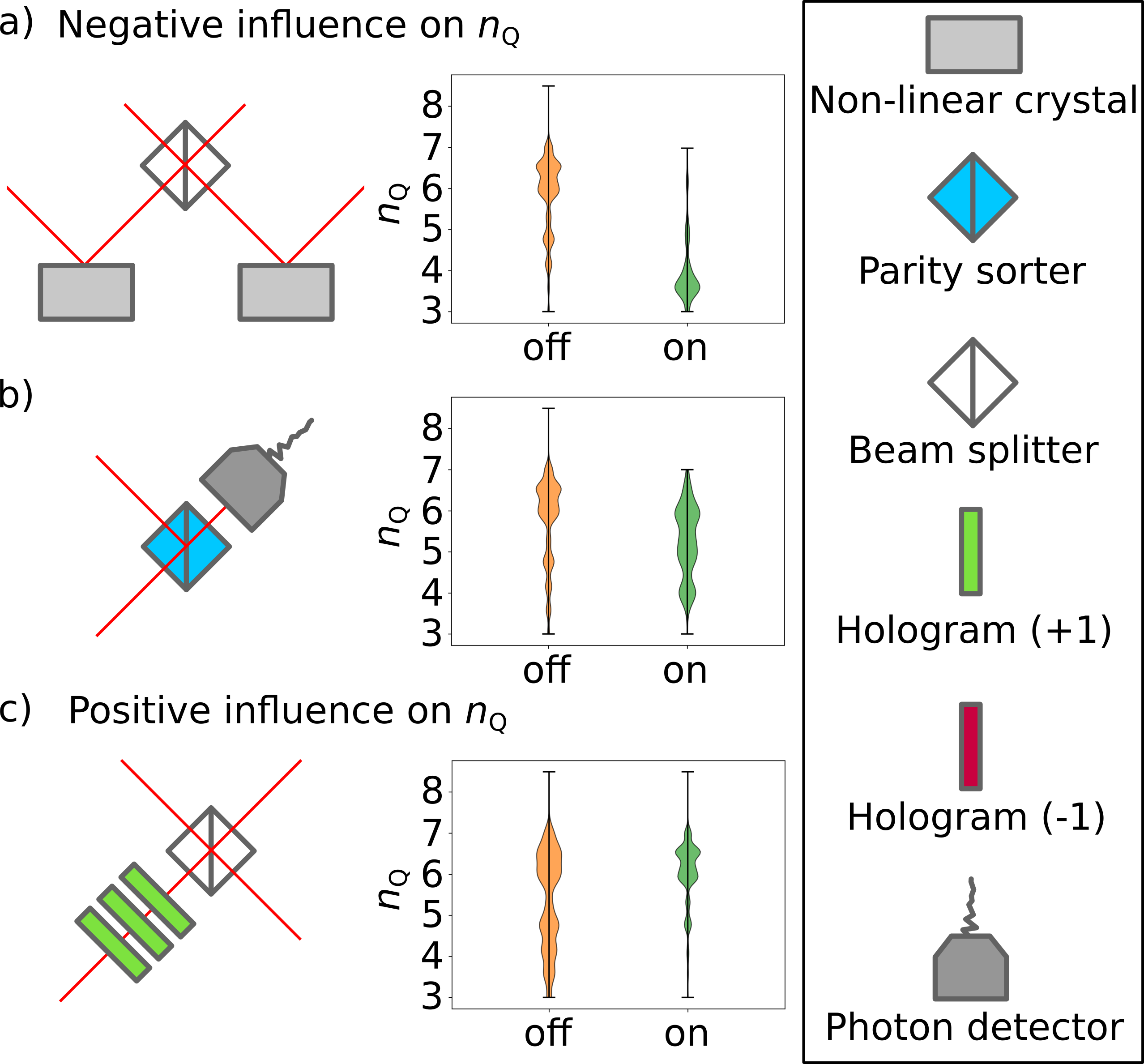}
    \caption{\textbf{Hypotheses about quantum optical experiments.} Experimental substructures leading to a decrease in the overall size of the Hilbert space of involved qubits ($n_Q$) are shown in a) while substructures with positive influence are shown in b).}
    \label{fig:quantum1}
\end{figure}

\begin{figure*}[hbt!]
    \centering
    \includegraphics[width=0.9\textwidth]{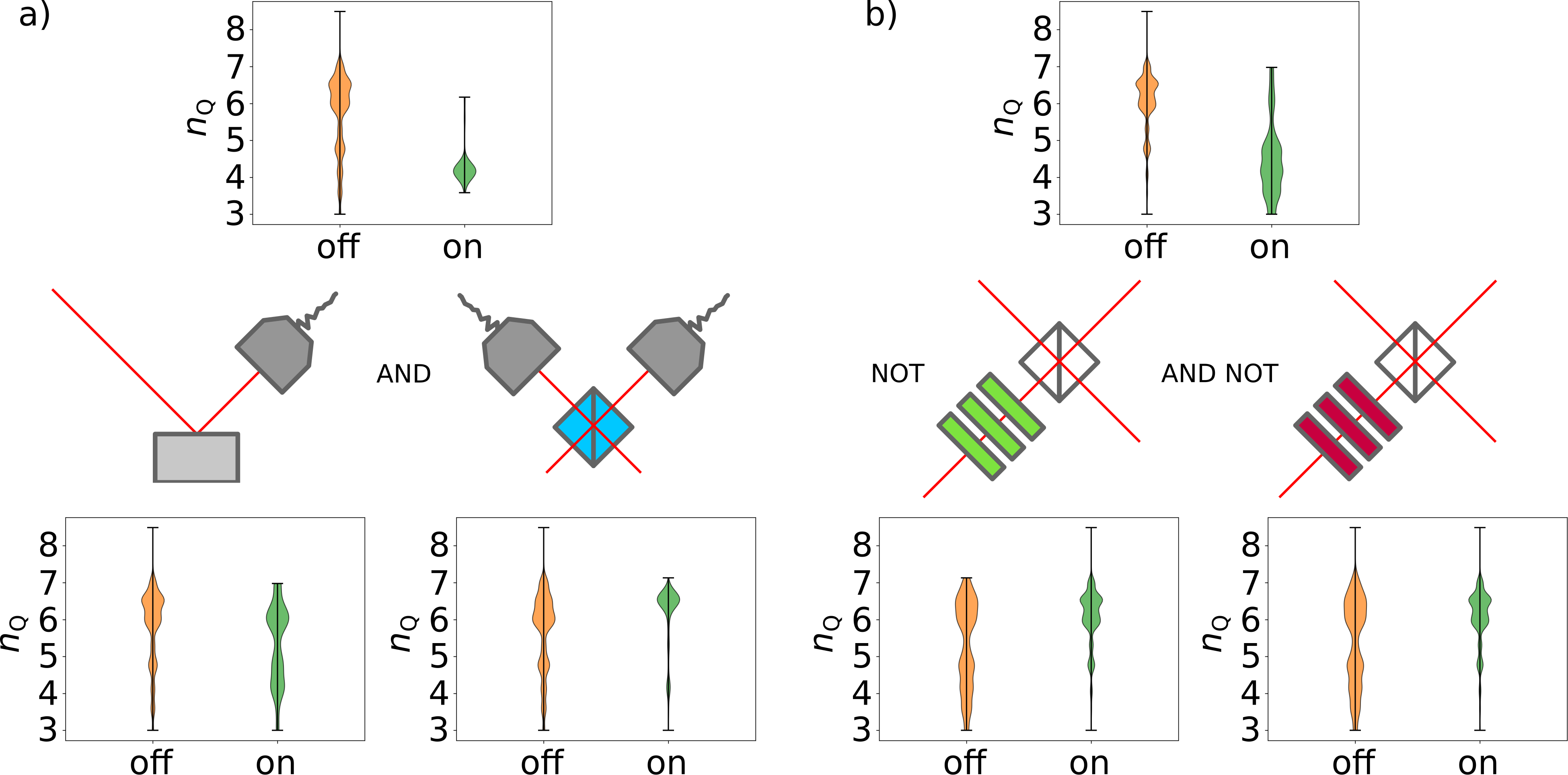}
    \caption{\textbf{Logically combined hypotheses about quantum optical experiments.} Combining single-subgraph hypotheses with logical operations leads to intuitively interpretable relations, which is illustrated here with two examples. The upper panel shows the logically combined feature and its correlation with $n_{\text{Q}}$, while the lower panels show the correlations of the isolated subgraphs.
}
    \label{fig:quantum2}
\end{figure*}

A key challenge lies in the design of experiments which creates certain desired quantum systems. The difficulty arises from counter-intuitive quantum phenomena, which raises the question of whether human intuition is the best way to design new experiments. Several studies have therefore developed automated and machine-learning augmented approaches for the design of experiments \cite{krenn2016automated, knott2016search,wallnofer2019machine,zhan2020experimental,krenn2020computer,krenn2020conceptual}. The goal in our approach is to tackle this challenge in a completely different way, namely by improving the scientist's intuition about these systems.\\

Specifically, we are investigating optical setups with three-photon entanglement in high dimensions, using a fourth photon as a trigger. The experimental setups can be represented as graphs where vertices represent optical elements, and edges correspond to the photon paths connecting these elements. Analogously to chemical elements, the optical elements can have one to four connections. For example, a beam splitter has four input-output modes, while a detector has only one input. As a measure of entanglement, we use the overall size of the involved Hilbert space in terms of involved qubits, $n_{\text{Q}} = \log_2(d_1 d_2 d_3)$, where $d_i$ stands for the rank of density matrix after tracing out photon $i$ \cite{huber2013structure,huber2013entropy}.\\

We used the same fingerprint-based graph representation as in Section \ref{chp:subsecChem} and trained a Gradient Boosting Regression model to predict $n_{\text{Q}}$. Using the algorithm outlined in Figure \ref{fig:overview}, we form a list of hypotheses of subgraphs features that influence $n_{\text{Q}}$ most. This computer-generated list was analysed and interpreted by a domain expert.\\

The two features which influence $n_{\text{Q}}$ most negatively contradict the intuition in the field, see Fig. \ref{fig:quantum1}a/b and Fig. S2. Surprisingly, both of them represent subgraphs that are core elements of two experimental setups which have produced high-dimensional multipartite entanglement in the laboratory\cite{malik2016multi, erhard2018experimental}. Specifically, if the outputs of two nonlinear crystals (both crystals produce entangled photon pairs in the same 3-dimensional mode space) are connected directly via a beam splitter or interferometer, the entanglement of the resulting state is predicted to be comparably low. This can be interpreted in the following way: The photons from the two different crystals need to combine at some point, otherwise, they remain bi-separable. However, if they combine directly after their generation, the equal mode spaces mix in such a way that it is difficult to increase their dimensionality subsequently. It is therefore explicitly enlightening that several of the features that positively influence $n_{\text{Q}}$ correspond to elements which shift the entire mode space by plus or minus three before or after the beam splitters or nonlinear crystals. The insight for a human researcher now is to shift the mode space by three (as the local dimension is three), before combining photons from different nonlinear crystals to achieve a high $n_{\text{Q}}$. This leads to mode spaces of twice the original size and thereby increasing the probability for large overall entanglement dimensionalities.\\

A different feature which was used in the two experimentally demonstrations, but significantly negatively influences $n_{\text{Q}}$ is the following: One output of a nonlinear crystal is directly connected to the detector. For human designers, this leads to the convenient fact that it simplifies the initial state (as double-emissions from one crystal can be ignored in this case). However, the entanglement of this photon with the other two photons can never be larger than three (as the local mode space is three). A similar, negatively influencing feature is a certain interferometer, which sorts the parity of the involved modes, directly connected to a detector. This acts as a filter, thus reducing the mode space of the incoming photon by half, thereby reducing the overall possible entanglement significantly.\\

\textbf{Logically combined features:} We can logically combine graph features, as described in \ref{chp:method}, and find the most significant macro-features for quantum experiments. In Fig. \ref{fig:quantum2}a, two small sub-experiments are combined with a logical \textit{and}, {\it i.e.} the feature is the combination of both structures. Individually, the presence of the first feature has a negative influence on $n_{\text{Q}}$. The second feature, a parity sorter followed by two detectors, influences $n_{\text{Q}}$ positively. Surprisingly, their combination has a significant negative influence on $n_{\text{Q}}$ and can be seen as an almost sufficient condition for $n_{\text{Q}}\approx 4$ . This behaviour can be interpreted using the Klyshko advanced wave-picture for quantum correlations in quantum optics \cite{klyshko1988simple}. The detector after the photon pair creation heralds a specific quantum state in the other photonic path. If those photons deterministically split at the parity sorter, the ability to mix with the photons from the other input ports (thus from the other crystal) vanish. From this insight, the human designer can learn that a heralded single-photon should be combined in a probabilistic way with the photons of the other crystal, using beam splitters instead of parity sorters.\\

A second macro-feature, Fig. \ref{fig:quantum2}b, combines two insights that we gained in Fig. \ref{fig:quantum1}. The macro-feature in Fig. \ref{fig:quantum2}b shows that the absence of either three positive or three negative mode shifters in front of a beam splitter has a very negative impact on the $n_{\text{Q}}$. Thereby, the algorithm has discovered that both increasing or decreasing helps to have very positive influence on the final entanglement, and thereby suggests that one can be agnostic about the shift direction, and the importance lies in the actual increase of the local Hilbert space before the mixing. This features clearly shows how logical combinations can simplify the interpretation of scientific data.\\

\section{Conclusion and outlook}
We presented a data-driven machine learning workflow for automated generation and verification of hypotheses about observations in natural sciences. We presented examples from chemistry and physics, but our method is directly applicable to most applications, where structures can be represented as graphs, {\it e.g.} to DNA/RNA data in biology \cite{gerling2015dynamic,praetorius2017biotechnological}, chemical reaction networks \cite{temkin1996chemical,rappoport2014complex} or graphs in social sciences. In chemistry, the workflow “rediscovers” widely known relations regarding solubility and electronic properties of molecules (often referred to as chemical intuition). In physics, the algorithm discovers rules to generate highly entangled three-photon states in quantum optical experiments. These rules are interpretable by human experts in retrospect, yet not known or postulated before, and even contradicting some of the field’s current understanding. Finding such rules will not only help researchers to understand complex scientific relationships and thus design better experiments, but also reduce unavoidable and often undetectable bias generated by prior knowledge and anticipations.\\

\textbf{Hypothesis testing.}
In addition to automated hypothesis generation, protocols for testing of the postulated hypotheses would be beneficial. In case of the chemistry experiment, a possible hypothesis testing protocol would generate mutations of each molecule in the training set to test the hypotheses on molecules with similar representations, where (ideally) only the relevant feature is changed. In case of the quantum optical experiments, not all random mutations will lead to maximally entangled states between all photons, which is a requirement to compute the entanglement of the quantum state. We currently see two options for automated hypothesis verification both of which we are currently implementing. The first follows the same procedure of mutation and computation as in the chemistry experiment, with the caveat that only a small fraction of the mutations will lead to useful results, potentially making the procedure computationally costly. The second option is based on finding other experimental setups within the whole database that are as similar to the reference experiment as possible, with the exception of the feature that is currently analysed. This procedure is computationally costly as well but does not require new computations.\\

\section*{Acknowledgements}
P.F. acknowledges funding the European Union’s Horizon 2020 research and innovation programme under the Marie Sk{\l}odowska-Curie grant agreement no. 795206 (MolDesign). M.K. acknowledges support from the Austrian Science Fund (FWF) through the Erwin Schr\"odinger fellowship No. J4309.  I.T. acknowledges NSERC and performed work at the NRC under the auspices of the AI4D and MCF Programs. A.A.-G. thanks Anders G. Fr{\o}seth for his generous support. A.A.-G. acknowledges the generous support of Natural Resources Canada and the Canada 150 Research Chairs program.

\section*{Data availabilty}
The data that support the findings of this study are available upon request from the authors.

\bibliographystyle{unsrt}
\bibliography{refs}

\clearpage
\newpage
\widetext

\setcounter{section}{0}
\setcounter{figure}{0}
\renewcommand{\thepage}{S\arabic{page}} 
\renewcommand{\thesection}{S\arabic{section}}  
\renewcommand{\thetable}{S\arabic{table}}  
\renewcommand{\thefigure}{S\arabic{figure}}

\begin{center}
\textbf{\large Supplementary Information:}

\textbf{\large Scientific intuition inspired by machine learning generated hypotheses}
\end{center}

\section{Additional hypothesis about chemistry data sets}

\begin{figure}[hbt!]
    \centering
    \includegraphics[width=0.95\textwidth]{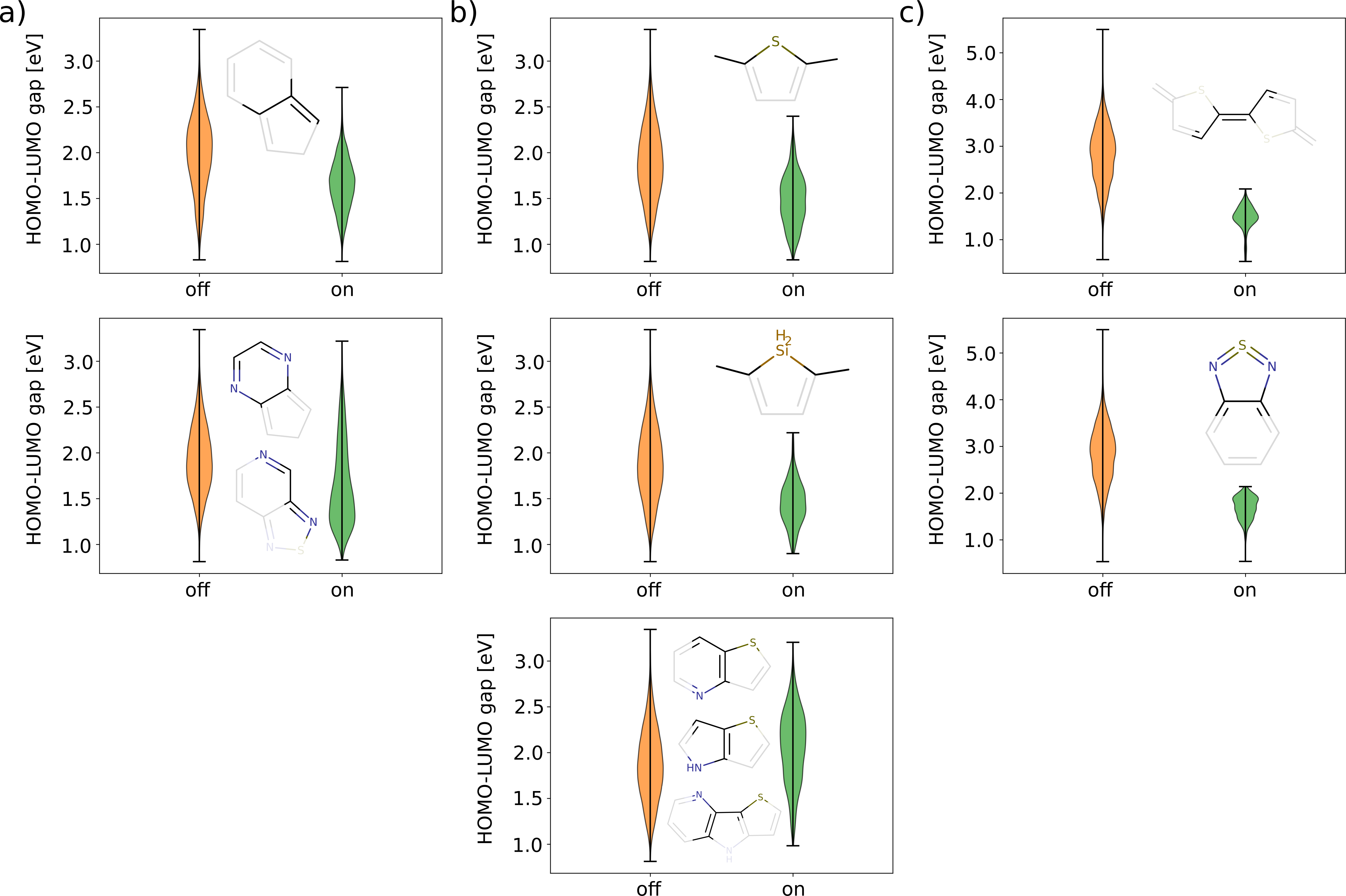}
    \caption{\textbf{HOMO-LUMO gaps in the Harvard Clean Energy data set \cite{hachmann2011harvard,lopez2016harvard} and a non-fullerene acceptor data set.\cite{lopez2017design}} (a) Agreeing with widely known rules of thumb, extended aromatic systems containing nitrogen heteroatoms are associated with reduced HOMO-LUMO gaps. (b) Thiophene but also more uncommon silole rings correlate with small HOMO-LUMO gaps. c) Thiophene rings bridged with double bonds ({\it i.e.} quinoid instead of aromatic systems) are found to decrease the HOMO-LUMO gap in the non-fullerene acceptor data set, a phenomenon that is studied and described in literature \cite{bredas1985relationship}. (Note the different scale in panel (c) compared to (a) and (b), due to differences in the data sets.)}
    \label{fig:homolumogapSI}
\end{figure}

Figure \ref{fig:homolumogapSI} shows additional features that influence the HOMO-LUMO gap of molecules in multiple data sets. Our workflow finds patterns that are commonly associated with a positive or negative influence on HOMO-LUMO gaps, but also patterns and groups such as silole rings.

\begin{figure}[hbt!]
    \centering
    \includegraphics[width=0.7\textwidth]{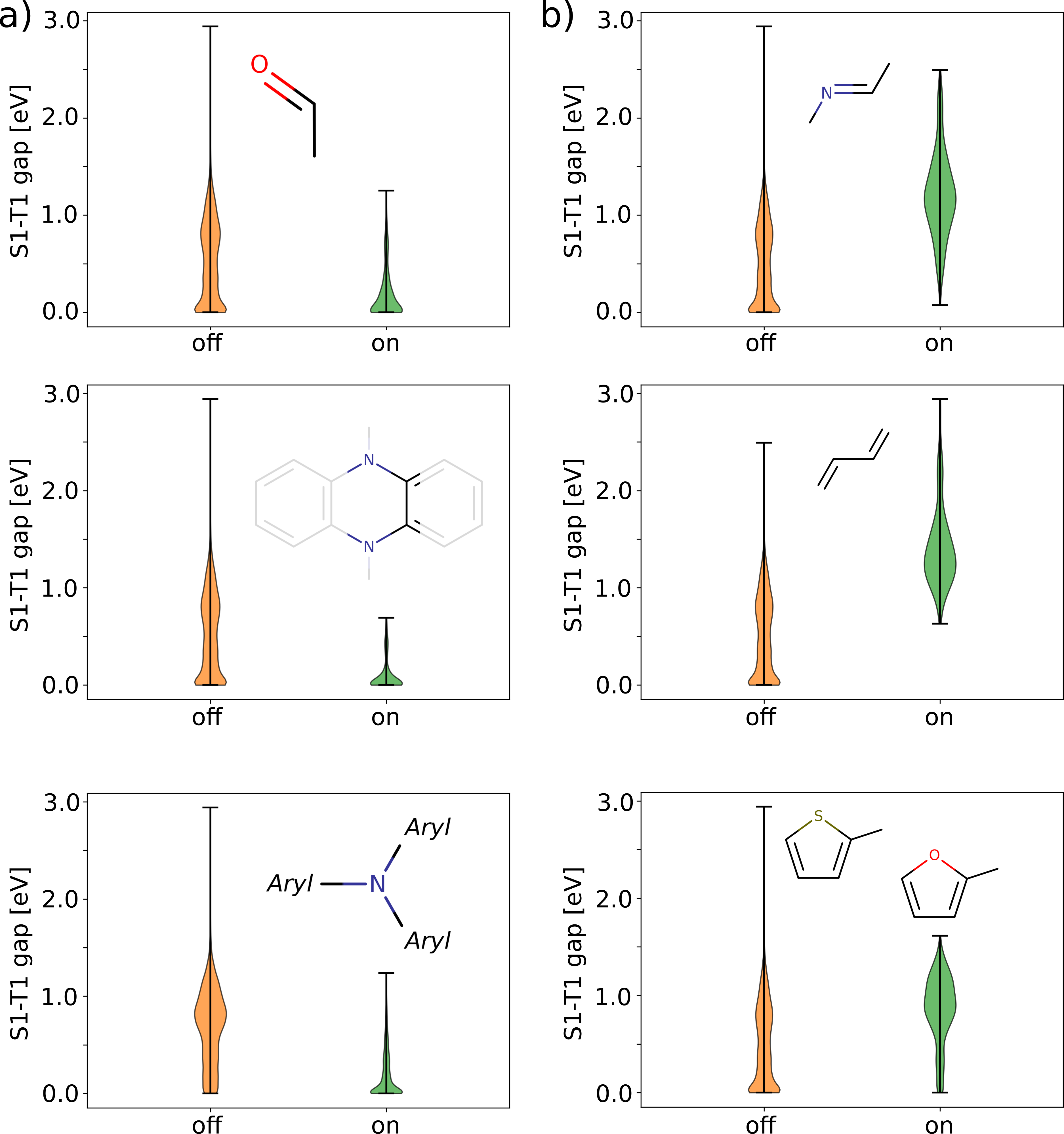}
    \caption{\textbf{TADF} (a) Groups with a negative influence on the singlet-triplet gap, (b) Groups with a positive influence on the singlet-triplet gap.}
    \label{fig:TADF_SI1}
\end{figure}

Figure \ref{fig:TADF_SI1} shows additional features that influence the singlet-triplet gap of TADF molecules. Figure \ref{fig:TADF_SI1}a shows groups with a negative influence on the singlet-triplet gap and Figure \ref{fig:TADF_SI1}b shows groups with a positive influence on the singlet-triplet gap.

\begin{figure}[hbt!]
    \centering
    \includegraphics[width=0.4\textwidth]{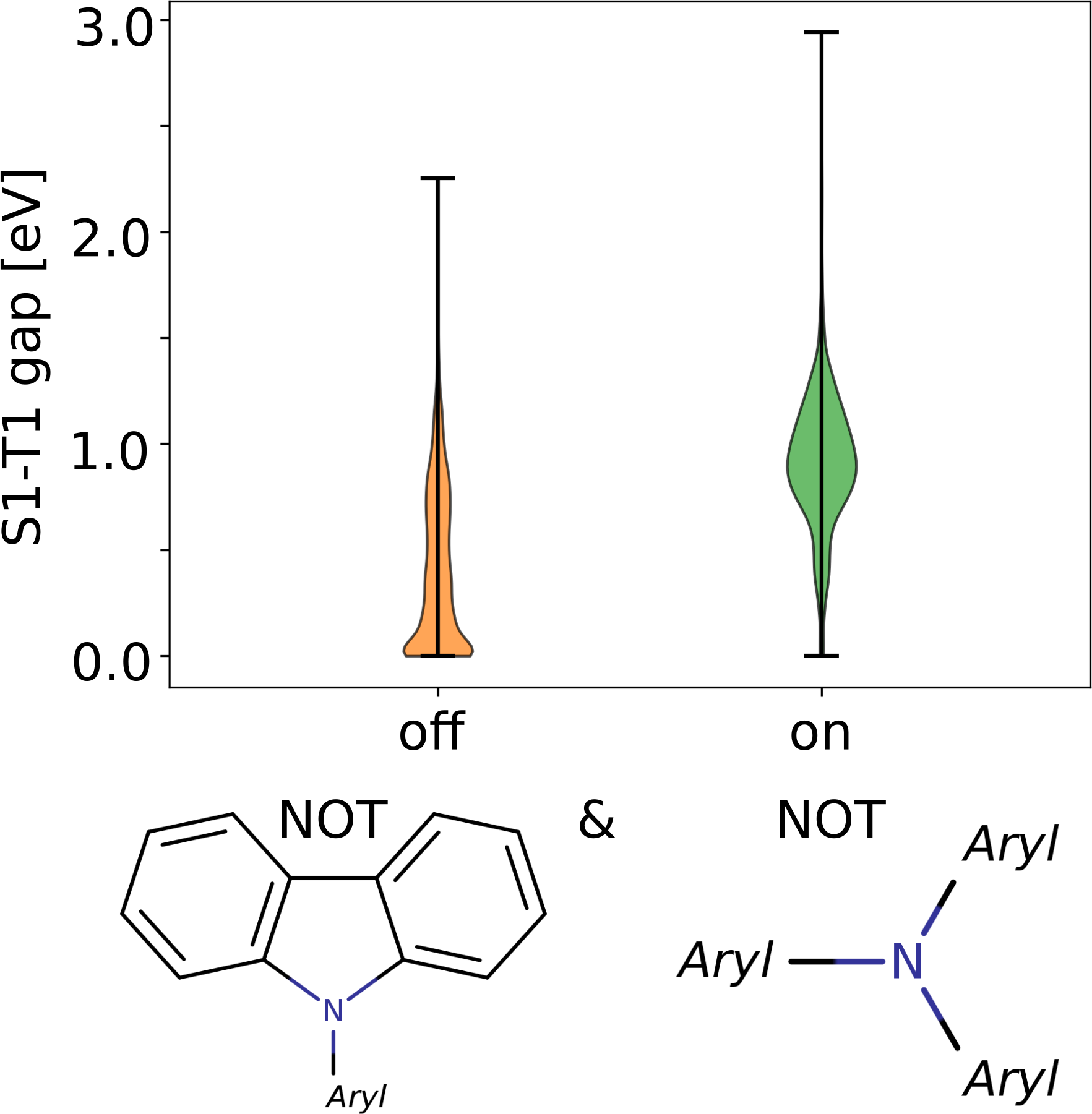}
    \caption{\textbf{TADF} The logical combination of carbazole and triarylamine groups indicate that the algorithm identified those two groups as required for small singlet-triplet gaps, because the absence of both groups leads to significantly increased singlet-triplet gaps.}
    \label{fig:TADF_SI2}
\end{figure}

Figure \ref{fig:TADF_SI2} shows an example of a easily interpretable, logically combined feature that influences the singlet-triplet gap of molecules.

\newpage
\newpage
\section{Additional hypothesis about quantum optical experiments}

\begin{figure}[hbt!]
    \centering
    \includegraphics[width=0.95\textwidth]{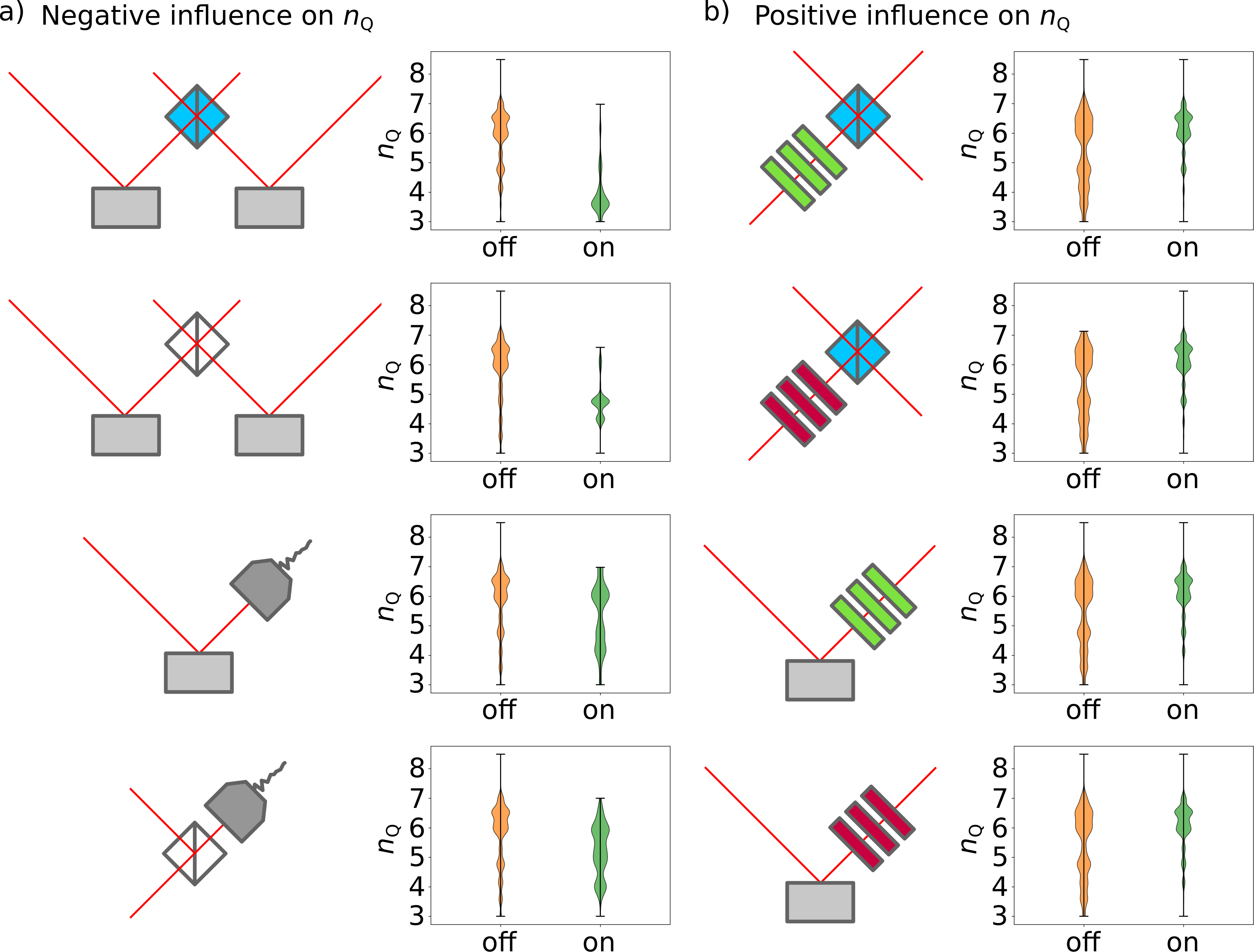}
    \caption{\textbf{Entanglement in quantum optical experiments} Subgraphs of the experimental setups that a) decrease and (b) increase the overall size of the involved Hilbert space $n_{\text{Q}}$}
    \label{fig:homolumogap}
\end{figure}

\end{document}